\pdfoutput=1
\documentclass[10pt,twocolumn,letterpaper]{article}
\usepackage{iccv}
\usepackage{times}
\usepackage{epsfig}
\usepackage{graphicx}
\usepackage{amsmath}
\usepackage{amssymb}
\usepackage[caption=false]{subfig}
\usepackage[accsupp]{axessibility}
\usepackage{setspace}
\usepackage{tabularx}

\usepackage[pagebackref=true,breaklinks=true,letterpaper=true,colorlinks,bookmarks=false]{hyperref}

\iccvfinalcopy 

\def\iccvPaperID{5961} 

\ificcvfinal\fi

\begin{document}

\title{Pre-trained Large Language Models for Financial Sentiment Analysis}

\author{Wei Luo$^{1}$\thanks{Equal contribution}, \ 
        Dihong Gong$^{2}$\footnotemark[1] \\ 
$^1$ Xi'an Jiaotong-Liverpool University, China \quad
$^2$ University of Florida, USA \\
{\tt\small luosting1234@163.com, gongdihong@gmail.com} 
}

\maketitle
\ificcvfinal\thispagestyle{empty}\fi
\begin{abstract}
\noindent 
Financial sentiment analysis refers to classifying financial text contents into sentiment categories (e.g. positive, negative, and neutral). In this paper, we focus on the classification of financial news title, which is a challenging task due to a lack of large amount of training samples. To overcome this difficulty, we propose to adapt the pretrained large language models (LLMs) \cite{radford2018improving, radford2019language, devlin2019bert} to solve this problem. The LLMs, which are trained from huge amount of text corpora, have an advantage in text understanding and can be effectively adapted to domain-specific task while requiring very few amount of training samples. In particular, we adapt the open-source Llama2-7B model (2023) with the supervised fine-tuning (SFT) technique \cite{touvron2023llama}. Experimental evaluation shows that even with the 7B model (which is relatively small for LLMs), our approach significantly outperforms the previous state-of-the-art algorithms. Data and code are available at \textcolor{red}{https://github.com/luosting/LLaMA-Financial-sentiment-analysis}.
\end{abstract}


\section{Introduction}
\noindent 
Sentiment analysis, a crucial facet of natural language processing (NLP), is pivotal in discerning opinions, attitudes, and emotions expressed in textual data. Within the financial domain, it serves as a critical tool for gauging market sentiments, customer feedback analysis, market research and competitive analysis brand monitoring and reputation management, financial trading and investment decisions, product development and marketing strategies, human resources and employee satisfaction and risk assessment \cite{liu_2020, 8187070, 10.5555/2021109.2021114, Cambria2015}.\\

\noindent 
The primary focus of this thesis revolves around polarity analysis, a process involving the categorization of textual content into positive, negative, or neutral sentiments within a defined domain. This undertaking presents two significant challenges: 1) Advanced classification techniques reliant on neural networks demand extensive labeled datasets. However, the labeling process for financial text excerpts necessitates substantial expertise, resulting in high costs. 2) Conventional sentiment analysis models trained on diverse corpora lack suitability for this task. Financial texts exhibit specialized language patterns, characterized by distinct vocabularies and a propensity for vague expressions rather than readily identifiable negative or positive terminologies. \\

\noindent 
Large language models represent a class of deep learning models possessing formidable natural language processing (NLP) capabilities. These models are trained on extensive corpora of text data, enabling them to effectively comprehend and generate natural language text. State-of-the-art large pre-trained language models (such as BERT, GPT) have exhibited remarkable performance improvements across a multitude of NLP tasks \cite{radford2019language, 10.5555/2021109.2021114}. They leverage transfer learning by pre-training on vast amounts of text data to acquire language knowledge and subsequently fine-tune on specific tasks, achieving impressive performance.\\

\noindent 
The evolution of sentiment analysis has been catalyzed by the emergence of large language models, a breakthrough in NLP. They possess pre-trained knowledge from extensive text corpora and excel at capturing complicate linguistic nuances, contextual comprehension, and semantic associations. In the context of finance, these large language models demonstrate exceptional proficiency in decoding financial texts, encompassing complex terminologies, domain-specific jargon, and subtle sentiment expressions. Their architecture, equipped with attention mechanisms and transformer-based structures, enables them to identify fine-grained sentiment variations, crucial for understanding market sentiments and investor behavior. \\

\noindent 
The integration of these advanced models into sentiment analysis methodologies augments the comprehension of financial data sources, including market reports, investment analyses, and customer reviews. Their ability to discern nuanced shifts in sentiment, anticipate market trends, and identify latent patterns elevates the precision and depth of sentiment analysis within financial contexts.\\

\noindent 
Previous studies by Agarwal and Mittal (2016), Guo et al. (2016), Araque et al. (2017), Devlin et al. (2018), and Dogu (2019) have significantly contributed to sentiment analysis methodologies and the utilization of machine learning and deep learning techniques within the financial domain. Their findings elucidate the advancements in sentiment analysis techniques and the potential applications of large language models in financial sentiment assessment. These researches apply the method of extracting features and then classifying them to achieve their intended goals, in other words, they use BERT-like method to cover it. However, unlike these researches, the aim of our research is fine-tuning LLaMA2-7B model using GPT-like method.

\noindent 
In this paper, we demonstrate the capability of LLMs for solving the financial sentiment analysis problem. The study is based on the Financial PhraseBank created by Malo et al. (2014)  dataset \cite{https://doi.org/10.1002/asi.23062}. The main contributions of this paper are summarized as follows.

\begin{itemize}
    \item We investigate with different methods of using few-shots, further pretraining, and SFT based on the Llama2-7B model.
    \item We achieve the state-of-the-art on PhraseBank financial sentiment analysis benchmark.
\end{itemize}

\section{Related Work}
\subsection{Previous work}
\noindent 
Sentiment analysis in the financial domain has attracted the attention of several scholars in recent years. Kraus and Feuerriegel (2017) explored the application of machine learning techniques for sentiment analysis by introducing LSTM, emphasizing the importance of feature engineering and model selection in achieving accurate sentiment classification \cite{Kraus_2017}. This study laid the groundwork for subsequent research endeavors aimed at enhancing sentiment analysis methodologies within financial contexts. In the study conducted by Sohangir et al. (2018) \cite{Sohangir2018BigDD}, various generic neural network architectures were employed on a dataset sourced from StockTwits. Their investigation revealed that the CNN emerged as the most effective neural network architecture among the models evaluated. Lutz et al. (2018) \cite{lutz2018sentencelevel}, alternatively, adopted a doc2vec approach to create sentence embeddings specific to company ad-hoc announcements. They further applied multi-instance learning techniques to predict stock market outcomes based on these embeddings. Maia et al. (2018) , on the other hand, utilized a combination of text simplification techniques alongside LSTM networks \cite{8334488}. Their objective was to classify a set of sentences extracted from financial news based on sentiment analysis. Their approach yielded state-of-the-art results for the Financial PhraseBank, a dataset also referenced in this thesis. However, these works fail to address the issue of incurring additional costs for labeling sample data, particularly when the available labeled data is insufficient to adequately fulfill the training requirements within such specific domain. 

\noindent 
Despite these limitations, recent advancements in deep learning have facilitated the utilization of large language models like BERT, which represents one of the variants of the Transformer model. Devlin et al. (2019) introduced BERT, a pre-trained deep bidirectional transformer model, revolutionizing language understanding by using mask \cite{devlin2019bert}. This breakthrough has significantly contributed to advancements in natural language understanding and sentiment analysis. In addition, the combination of deep learning language models and supervised fine-tuning (SFT) has emerged as a compelling approach in various domains, demonstrating enhanced capabilities in understanding and generating contextually relevant information. Consequently, Dogu (2019) integrated BERT with SFT, creating the FinBERT model specifically tailored for financial sentiment analysis. FINBERT utilizes a training corpus consisting of datasets such as TRC2-financial, Financial PhraseBank, and FiQA Sentiment and several optimization strategies including slanted triangular learning rates, discriminative fine-tuning and gradual unfreezing \cite{araci2019finbert}. Through implementing the aforementioned methods, the principal contribution lies in a substantial enhancement of deep learning network accuracy specifically tailored for financial sentiment analysis. As discussed by Yang et al. (2020), it demonstrated impressive capabilities in financial sentiment analysis and stock prediction tasks. Their research highlighted the superior performance of FinBERT when compared to generic sentiment analysis models in deciphering sentiment nuances within financial texts \cite{yang2020finbert}. These studies underscore the growing trend of customizing language models, like FinBERT, to cater to the intricate demands of the financial domain, showcasing their prowess in decoding financial jargon and capturing subtle sentiment shifts.

\noindent 
Despite the commendable performance of FINBERT in financial sentiment analysis, our research project aims to explore an alternative approach distinct from BERT for this purpose. Specifically, we are contemplating the utilization of the GPT (Generative Pre-trained Transformer) model. Based on the paper of Yang et al. (2023) and Kheiri et al. (2023), the GPT model differs significantly from BERT in its pre-training methodology and architectural design, primarily emphasizing its generative modeling capabilities. In contrast to BERT’ s bidirectional encoding, the GPT model adopts a unidirectional architecture, focusing more on context generation and coherence \cite{10.1007/978-3-031-40292-0_19, kheiri2023sentimentgpt}. We hypothesize that the GPT model, owing to its generative nature, might offer a distinctive perspective and advantages in the realm of financial sentiment analysis compared to conventional models. While acknowledging the remarkable performance of FINBERT, our intent is to explore the application of GPT in financial sentiment analysis to present additional choices and possibilities in this domain. This research direction aims to delve deeper into the potential of GPT in financial sentiment analysis, seeking novel insights and solutions.

\subsection{LLaMA model}
\noindent 
LLaMA, a derivative of the GPT model modified by Hugo et al. (2020), features a substantial reduction in parameter count compared to GPT \cite{touvron2023llama}. The maximum parameter count among its four versions is 65 billion, considerably fewer than GPT-3’ s 175 billion parameters. Its lowest-parameter version, consisting of 7 billion parameters, suggests the potential for efficient deployment even on units with slightly inferior hardware due to reduced computational demands.

\noindent
LLaMA applies open-source corpus consisting of the text from English CommonCrawl, C4, Github, Wikipedia, Gutenberg and Books3, ArXiv, Stack Exchange for training and this corpus contains 1.4T tokens after tokenization. During the training phase, special emphasis is placed on gathering high-quality supervised fine-tuning (SFT) data examples to maintain the model’ s consistency in conversational instruction-related tasks. These examples effectively guide the model's fine-tuning process, thereby enhancing its performance and generalizability for specific tasks. Each training sample comprises a prompt and a response, delineated by specific separators. The use of autoregressive objectives involves setting the token loss of user prompts to zero, allowing backpropagation solely on answer tokens. This strategy aims to enhance the accuracy and coherence of the generated responses. In terms of model rewards, the research team uses a reward model to evaluate the generated responses of the model and generates a scalar score to measure the quality of generation. The initialization of the reward model uses pre-trained chat model checkpoints and replaces the original next token prediction classifier with a scalar reward value regressor. During the training process, a binary sorting loss function with marginal constraints is adopted to improve the accuracy of the reward model.

\section{Method}
\noindent 
In this section, we present our algorithms for using the pretrained LLaMA-7B model for financial sentiment analysis. 

\subsection{LLM Few-shot Prediction} \label{sec-fewshots}
\noindent 
In the context of few-shot prediction utilizing large language models, the process involves crafting appropriate prompts to guide the model in generating the desired results. For the task of financial sentiment analysis, the prompt is formulated as follows:

\begin{minipage}{0.4\textwidth}
{\it
We want to perform the sentiment analsysis for financial news according to their titles.
You are asked to choose the most suitable sentiment from ("postive", "negative", "neutral") with signle-choice questions.
Please follow these examples to answer the question.
\newline
\newline
News Title: About Nokia Nokia is a pioneer in mobile telecommunications and the world's leading maker of mobile devices.\newline
Choices: A) positive. B) negative. C) neutral.\newline
Answer:A
\newline
\newline
News Title: Most of the permanent layoffs will be in the plywood and sawn timber sectors of the Finnish company's operations at several domestic mills, where earlier this year it temporarily laid off some 1,200 workers to save costs.\newline
Choices: A) positive. B) negative. C) neutral.\newline
Answer:B
\newline
\newline
News Title: The transaction is planned to be financed with a EUR40m market-based loan granted by Standard Chartered Bank Hong Kong.\newline
Choices: A) positive. B) negative. C) neutral.\newline
Answer:C
\newline
\newline
{\bf
News Title: news title here \newline
Choices: A) positive. B) negative. C) neutral. \newline
Answer:}
}
\newline
\end{minipage}

\noindent 
In the provided prompt template, a 3-shot format is utilized, consisting of 3 example questions and their corresponding answers. This format is followed by the news title to be classified, highlighted in bold text. By substituting the appropriate news title into the prompt, the model is anticipated to generate an answer, selecting one of the three options: A, B, or C. Consequently, the model effectively accomplishes the classification task as intended.

\subsection{Supervised Fine-Tuning}
\label{sec-sft}
\noindent 
Classification accuracy of LLMs can be further improved with SFT on specific downstream task. In this section, we present our design of fine-tuning the LLaMA 7B base model to adapt the pretrained model to our financial sentiment analysis task. We follow the same paradigm of formulating the problem as a single-choice question among \{"positive", "negative", "neutral"\} as described in Section~\ref{sec-fewshots}. 

\noindent 
Suppose the training data $D$ consisting of $N$ pairs of questions and the corresponding answers is defined as follows:
\begin{equation}
    D=\{(Q_n, A_n)|n=1,...,N\}
\end{equation}
where $Q_n$ is a financial news title, and $A_n$ is the annotated sentiment label. Given $D$ we perform tokenization for the questions and answers, respectively. The tokenized output of each question consists of a sequence of tokens with variable lengths, whereas the tokenized output of an answer comprises a single token. Consequently, the set $D$ represents the tokenized data derived from $T$:
\begin{equation}
    T=\{(\vec{q_n}, a_n)|n=1,...,N\}
\end{equation}
where $q_n=\text{tokenize}(\text{prompt}(Q_n))$, the $\text{tokenize}(\cdot)$ means tokenization operation, and $\text{prompt}(\cdot)$ means creating prompt text using the previously described prompt template. Then we organize the training sequence $X$ that is subsequently fed into the transformers as follows:
\begin{equation}
    X=|\text{BOS}|\vec{q_1}|\text{EOS}|a_1|...|\text{BOS}|\vec{q_K}|\text{EOS}|a_K|
\end{equation}
where $K$ is the number of question-answer pairs loaded into a single training sequence, and BOS is the begin of sequence token while EOS is the end of sequence token. The operation $P(\cdot)$ means creating prompt tokens

\noindent
The prediction label $Y$ is organized as follows:
\begin{equation}
    Y=|\vec{E_1}|E|a_1|...|E|\vec{E_K}|E|a_K|E|
\end{equation}
where $E=-1$ indicates the corresponding logit is not counted towards the cross entropy loss, and $\vec{E_k}$ is a vector of $-1$ whose length is the same as the $q_k$. As a result, we only count the loss of the answer parts $a_1,...,a_K$.

\noindent
To prevent mutual interference for different training pairs, we employ the block-diagonal attention mask, so that each question-answer pair sees information within their own context. In particular, when predicting $a_k$, we only uses the information of $q_k$, not using any preceding information of $k-1,..,0$.

\subsection{Sentiment Analysis with Classification Head}
\label{sec-scwop}
\noindent 
In this section, we model the sentiment analysis problem as a multi-class classification problem, rather than a answer generation problem as described in section~\ref{sec-sft}. One potential advantage of modeling the problem as a classification rather than generation problem is that it provides classification probability that is useful for many practical applications. We organize the training sequence $X$ that is subsequently fed into the transformers as follows:
\begin{equation}
    X=|\text{BOS}|\vec{q_1}|\text{EOS}|
\end{equation}
Note that we only load one sample per microbatch. The corresponding classification label is $Y \in \{0,1,2\}$ where $0$ means "positive", $1$ means "negative" and $2$ means "neutral".

\section{Experiments}
\label{sec:experiments}

\subsection{Implementation Details}
\noindent
The pretrained model we employ in this research is LLaMA-7B version 2 without SFT. The model is trained with 5 epochs of the training data, with learning rate starting from $3e^{-5}$ and reduces to $3e^{-6}$ with cosine annealing schedule. The gradient clip is $1.0$ to prevent gradient explode. The micro batch size is set to $4$ with max sequence length $1024$. The BFloat16 is used at training to prevent data overflow or underflow. While at the testing stage, Float16 is used for optimal numerical precision. The model is trained with data parallel paradigm with world size of $4$ and each GPU has RAM of 80GB.

\subsection{Dataset}
\noindent
\textit{Financial PhraseBank.} In this paper, following the article \cite{yang2020finbert}, we deploy a publicly available data set distributed by Malo et al. (2014) \cite{https://doi.org/10.1002/asi.23062} for our study. The dataset contains 4845 english sentences randomly extracted from financial news found on LexisNexis database. For fair comparison, we follow the same data segmentation split from \cite{yang2020finbert} to divide data into three sets: training, validation and testing. In particular, $20\%$ of the data is held out for testing, while the reset is used for training and model validation. The performance of algorithms is measured with classification accuracy. 

\subsection{Ablation Study}

\begin{table}[t]
\label{tab-result1}
\centering
\begin{spacing}{1.25}
\begin{tabular}{|c|c|c|c|}
\hline
{\bf Methods} & {Base} & {SFT} & {ClassHead} \\
\hline
{\bf Accuracy} & 0.68 & 0.90 & 0.90  \\
\hline
\end{tabular}
\end{spacing}
\caption{Ablation study for comparing accuracy of three methods: Base, SFT and ClassHead.}
\end{table}

\noindent
In this section, we conduct ablation study to investigate the performance of different components of our proposed method. We have three versions of methods: the base method as described in section~\ref{sec-sft}, SFT method as described in section~\ref{sec-fewshots}, and ClassHead method as described in section~\ref{sec-scwop}. The base method leverages the capability of pretrained LLMs, while the SFT method further improves this capablity by fune-tuning the model use task-specific training data, and finally the ClassHead method models the problem as classification problem by adding a classification head at the output layer. The results of these three methods are show in Table~1. According to these results, we can see that the SFT version improves from the base version by a clear margin, which confirms that our SFT method is effective in improving the classification accuracy of the financial news titles. While the ClassHead method performs equally with the SFT method, which indicates that the generation and classification modeling method have the same accuracy. However, the ClassHead method has an extra advantage of outputing the classification confidence score in addition to the class label, which is useful for many practical applications.

\subsection{Experimental Results}

\begin{table}[t]
\label{tab-result2}
\centering
\begin{spacing}{1.25}
\begin{tabular}{|c|c|}
\hline
{\bf Methods} & {\bf Accuracy} \\
\hline
LSTM\cite{araci2019finbert, schmidhuber1997long} & 0.71 \\
LSTM with ELMo\cite{araci2019finbert, peters-etal-2018-deep} & 0.75 \\
ULMFit\cite{araci2019finbert, howard2018universal} & 0.83 \\
LPS\cite{malo2014good} & 0.71 \\
HSC\cite{krishnamoorthy2018sentiment} & 0.71 \\
FinBERT\cite{araci2019finbert} & 0.86 \\
\hline
\textbf{Ours(SFT)} & \textbf{0.90} \\
\textbf{Ours(ClassHead)} & \textbf{0.90} \\
\hline
\end{tabular}
\end{spacing}
\caption{Benchmark results for financial news sentiment analysis.}
\end{table}

\noindent
In this section, we compare our result with those of the state-of-the-art methods. The details of the results are showed in Table~2. The results show that our method has improved the current state-of-the-art accuracy from 0.86 to 0.9, which is a big improvement. This reveals that pretrained LLMs are much more powerful for text understanding and classification than BERT-based methods. 

\section{Conclusion}
\noindent
In this paper, we explored the potentials of using LLMs for the financial sentiment analysis. We performed a systematically empirical analysis and provided novel insights on how to efficiently utilize the LLMs to improve the classification accuracy. Specifically, the few-shots only method can achieve relatively decent accuracy, while the further pretraining doesn't provide a noticeable improvement over its baseline. Finally, our SFT algorithm significantly improves over the baseline method and achieve a new state-of-the-art performance. Future work includes using LLMs with larger number of parameters (e.g. LLaMA2-70B).

{\small

\bibliographystyle{unsrt}
\bibliography{main}
}

\end{document}